\documentclass[10pt,twocolumn,letterpaper]{article}

\usepackage{iccv}
\usepackage{times}
\usepackage{epsfig}
\usepackage{graphicx}
\usepackage{amsmath}
\usepackage{amssymb}
\usepackage{ulem}
\usepackage[dvipsnames,table]{xcolor}
\usepackage{graphicx}
\usepackage{amsmath}
\usepackage{amssymb}
\usepackage{booktabs}
\usepackage{xcolor}
\usepackage{color}
\usepackage{pifont}
\usepackage{float}
\usepackage[caption=false]{subfig}
\usepackage{bm}
\usepackage[T1]{fontenc}
\usepackage[dvipsnames]{xcolor}
\usepackage[accsupp]{axessibility} 
\definecolor{RowColor}{rgb}{0.95, 0.95, 1}
\definecolor{citegrey}{HTML}{75878a}

\usepackage[pagebackref=true,breaklinks=true,letterpaper=true,colorlinks,bookmarks=false]{hyperref}
\newcommand{\cmark}{\textcolor{green}{\ding{51}}}%
\newcommand{\xmark}{\textcolor{red}{\ding{55}}}%
\newcommand{\tablestyle}[2]{\setlength{\tabcolsep}{#1}\renewcommand{\arraystretch}{#2}\centering\footnotesize}
\iccvfinalcopy 

\newcommand{\RDB}[1]{\textcolor{black}{#1}}

\def\iccvPaperID{6260} 

\ificcvfinal\fi
\makeatletter
\def\thanks#1{\protected@xdef\@thanks{\@thanks
        \protect\footnotetext{#1}}}
\makeatother
\begin{document}

\title{Sample-adaptive Augmentation for Point Cloud Recognition \\ Against Real-world Corruptions}

\author{
\textbf{Jie Wang$^{1}$, Lihe Ding$^{1}$, Tingfa Xu$^{1,\dag}$, Shaocong Dong$^{1}$, Xinli Xu$^{1}$, Long Bai$^{2}$, Jianan Li$^{1,\dag}$}
\\
$^{1}$~Beijing Institute of Technology~~~~~ $^{2}$~The Chinese University of Hong Kong
\\
\thanks{
$^{\dag}$Correspondence to: Jianan Li and Tingfa Xu.\\
}
\tt\small \url{https://github.com/Roywangj/AdaptPoint}
}

\maketitle
\ificcvfinal\thispagestyle{empty}\fi

\begin{abstract}
Robust 3D perception under corruption has become an essential task for the realm of 3D vision. While current data augmentation techniques usually perform random transformations on all point cloud objects in an offline way and ignore the structure of the samples, resulting in over-or-under enhancement. In this work, we propose an alternative to make sample-adaptive transformations based on the structure of the sample to cope with potential corruption via an auto-augmentation framework, named as AdaptPoint.  Specially, we leverage a  imitator, consisting of a Deformation Controller and a Mask Controller, respectively in charge of predicting deformation parameters and producing a per-point mask, based on the intrinsic structural information of the input point cloud, and then conduct corruption simulations on top. Then a discriminator is utilized to prevent the generation of excessive corruption that deviates from the original data distribution. In addition, a perception-guidance feedback mechanism is incorporated to guide the generation of samples with appropriate difficulty level. Furthermore, to address the paucity of real-world corrupted point cloud, we also introduce a new dataset ScanObjectNN-C,
\RDB{that exhibits greater similarity to actual data in real-world environments, especially when contrasted with preceding CAD datasets.} Experiments show that our method achieves state-of-the-art results on multiple corruption benchmarks, including ModelNet-C, our ScanObjectNN-C, and ShapeNet-C.


\end{abstract}

\vspace{-5mm}
\section{Introduction}
\label{sec:intro}
3D vision has gained attention for its potential applications in robotics and autonomous driving. Existing methods~\cite{qi2017pointnet, qi2017pointnet++, wang2021papooling, qianpointnext} primarily focused on clean data~\cite{modelnet40, chang2015shapenet} with few corruptions, whereas practical applications often entail numerous corruptions. Consequently, these models exhibit poor performance in such scenarios. As such, improving the robustness of point cloud models against corruption remains a crucial but challenging problem.

\begin{figure}[t!]
\centering
  \includegraphics[width=0.45\textwidth]{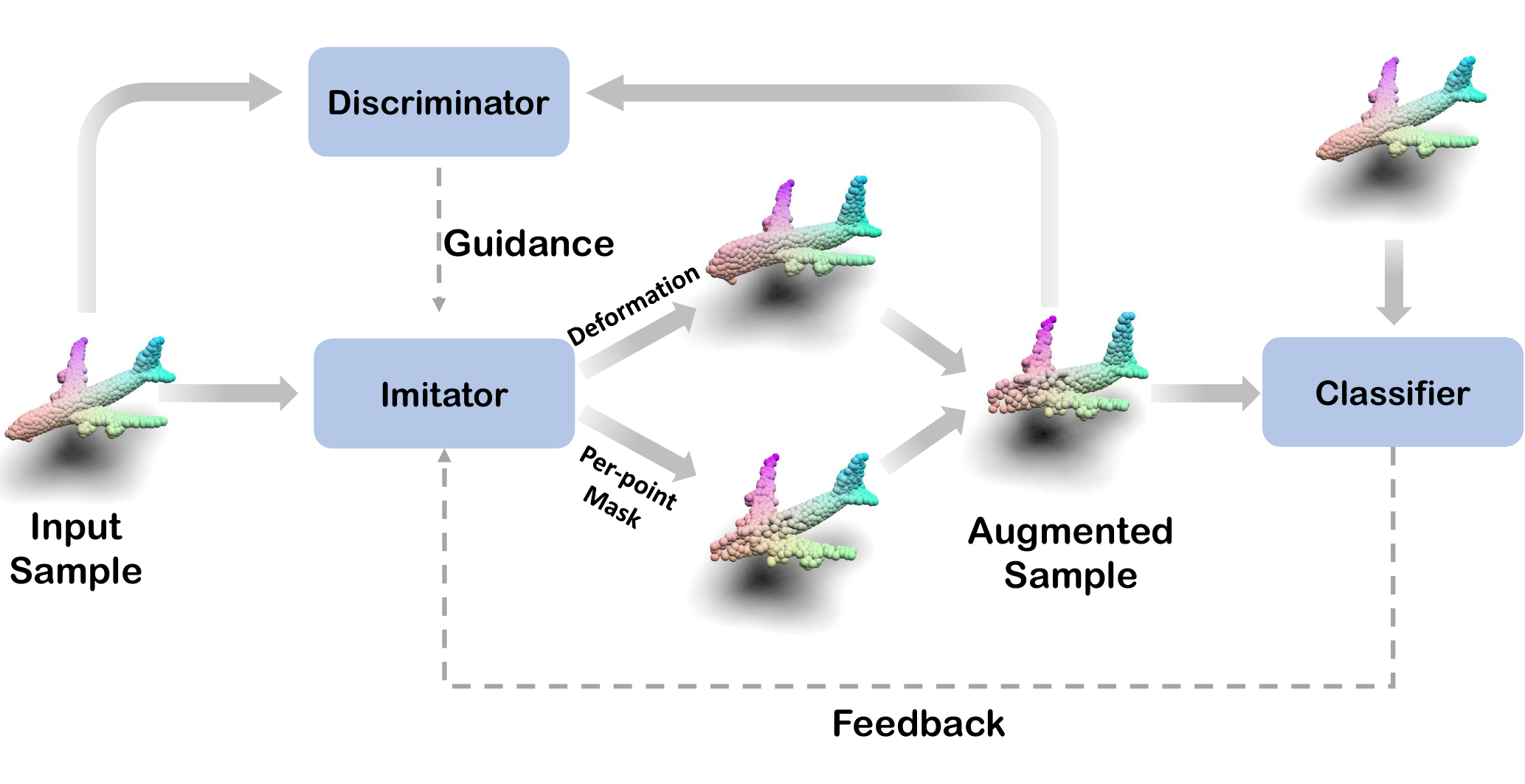}
  \caption{\textbf{Workflow of AdaptPoint}. The AdaptPoint framework involves a co-evolving strategy where the imitator, discriminator, and point cloud classifier are jointly trained. During the training process, the imitator is trained to transform clean data into realistic corruptions while receiving feedback and guidance from the classifier and discriminator, respectively. } 
  \label{AdaptPoint Workflow}
  \vspace{-6mm}
\end{figure}

To enhance the model's resistance to corruption, a frequently employed technique in prior studies is data augmentation~\cite{kim2021point,lee2021regularization,ren2022benchmarking}. Despite their effectiveness, these methods are subject to two common limitations. Firstly, these offline methods fail to consider the structure of the classifier, which may lead to the generation of samples that are not beneficial for enhancing model robustness. Additionally, they often overlook intrinsic features of the data and apply random transformations to all samples, resulting in augmentations that are either overly difficult or overly simplistic. 

To tackle the issues at hand, we introduce a new auto-augmentation framework, AdaptPoint, that is specifically tailored to generate augmented samples that consider the inherent structure of input point cloud. 
At the heart of our proposed method lies the sample-adaptive imitator network, which is designed to generate point cloud that are representative of real-world scenarios. It first extracts local geometric features from the input point cloud, resulting in a subset of points that encode these features. Subsequently, the imitator network is divided into two branches, namely the \textbf{Deformation Controller} and the \textbf{Mask Controller}.  The former module, is tasked with modifying the point cloud structure by incorporating the interactions between local regions and predicting the deformation parameters for specific local areas. On the other hand, the Mask Controller,  is responsible for selectively removing certain points by leveraging the overall structural information of the point cloud and generating a per-point mask.
Finally, the deformation and mask parameters are fed into a corruption simulator to generate a corrupted point cloud. The design of the deformation and mask mechanisms endows the imitator with the ability to simulate most real-world corruptions. 


The sample-adaptive imitator designed has an impressive capability to generate samples that closely resemble real-world corruptions.
The following problem to be considered is how to provide sufficient guidance to the imitator network to ensure: i) the generated samples conform to the real point cloud distribution; and ii) generating the samples with appropriate difficulty for classifier training. To achieve these objectives, we introduce a discriminator and a classifier. The discriminator is used to distinguish between the distributions of clean and augmented samples, thereby preventing the generation of excessive corruption that deviates from the original data distribution. On the other hand,
the classifier is employed to provide classification error feedback to guide the imitator network in generating samples with an appropriate level of difficulty. These three key components are integrated into the proposed AdaptPoint framework, as illustrated in Fig.~\ref{AdaptPoint Workflow}. By jointly optimizing the imitator with a point cloud discriminator and classifier, our framework progressively generates corrupted data, thereby enhancing the model's robustness against corruption.


Current datasets for point cloud corruption evaluation~\cite{ren2022benchmarking,ren2022pointcloud} are largely limited to CAD models, which fail to capture the variety of real-world corruptions encountered in practice and may hinder research progress in point cloud robustness. To bridge this gap, we present a new point cloud corruption dataset, namely ScanObjectNN-C, which is curated by collecting real-world point clouds from ScanObjectNN~\cite{uy2019revisiting} dataset, along with corresponding annotations and five types of corruptions: ``Jitter'',``Drop Global/Local'', ``Add Global/Local'', ``Scale'', and ``Rotate''~\cite{ren2022benchmarking}.


The effectiveness of AdaptPoint is validated through extensive experimentation on diverse datasets such as ModelNet-C~\cite{ren2022benchmarking}, ShapeNet-C\cite{ren2022pointcloud}, and our more challenging ScanObjectNN-C. Our proposed AdaptPoint achieves state-of-the-art results on these datasets, indicating its capability to alleviate the negative influence of data corruptions on point cloud perception performance.
Furthermore, our method achieves competitive results in defending point cloud attacks, indicating its generality and effectiveness in handling different types of corruption scenarios.

The main contributions of our work are as follows:
\begin{itemize}
    \vspace{-2mm}
    \item We present a new auto-augmentation framework, AdaptPoint, which is tailored for point cloud recognition in the presence of real-world corruptions.
    \vspace{-2mm}
    \item We propose a novel sample-adaptive imitator network that can simulate various types of realistic corruptions by leveraging the intrinsic characteristics of input.
    \vspace{-2mm}
    \item We construct a real-world point cloud corruption dataset ScanObjectNN-C, which can facilitate research on real-world point cloud corruption.
    \vspace{-2mm}
    \item We establish new state-of-the-art results on several corruption benchmarks.
\end{itemize}

\section{Related Work}
\noindent\textbf{Deep Learning on Point Cloud Corruption.}
Despite the efficacy of existing models in point cloud perception~\cite{qi2017pointnet, qi2017pointnet++, ma2021rethinking, qianpointnext, wu2019pointconv, xu2021paconv, wang2021papooling}, they remain vulnerable to corruption. Recent research has highlighted the increasing importance of mitigating point cloud corruption~\cite{ren2022benchmarking,ren2022pointcloud,kim2021point,lee2021regularization}. Kim et al.\cite{kim2021point} proposed PointWolf, which enhances point cloud model robustness using non-rigid deformation, while Lee et al.\cite{lee2021regularization} proposed RSMix, which generates mixed samples by rigidly transforming two point clouds. Ren et al.\cite{ren2022benchmarking} proposed WOLFMix, which combines PointWOLF and RSmix to further enhance model robustness. Additionally, Ren et al.\cite{ren2022pointcloud} proposed an attention-based backbone RPC for point cloud corruption. Although these methods have shown improvement in model robustness against corruption to some extent, they may not perform well on real-world datasets. Our proposed online auto-augmentation framework, AdaptPoint, provides a solution to this vulnerability.

\noindent\textbf{Data Augmentation on Point Cloud.}
Data augmentation has become widely accepted in the deep learning community for enhancing the generalization performance of deep neural networks.
However, traditional data augmentation techniques such as random scaling, rotation, and jitter have limitations in improving the capability of models for point cloud analysis. To overcome this, advanced data augmentation techniques have been proposed in recent studies, such as PointAugment~\cite{li2020pointaugment}, PointMixup~\cite{chen2020pointmixup}, and PointWOLF~\cite{kim2021point}. Nevertheless, PointMixup~\cite{chen2020pointmixup} and PointWOLF~\cite{kim2021point} primarily perform pre-defined transformations, and PointAugment~\cite{li2020pointaugment} only learns sample-wise global transformations while neglecting the significance of local geometry in point cloud. In this paper, we introduce a novel auto-augmentation framework for combating corruptions that can adjust augmentations according to both global and local features of point clouds, as well as the current capability of the model.

\begin{figure*}[!t]
\centering
\includegraphics[width=0.95\textwidth]{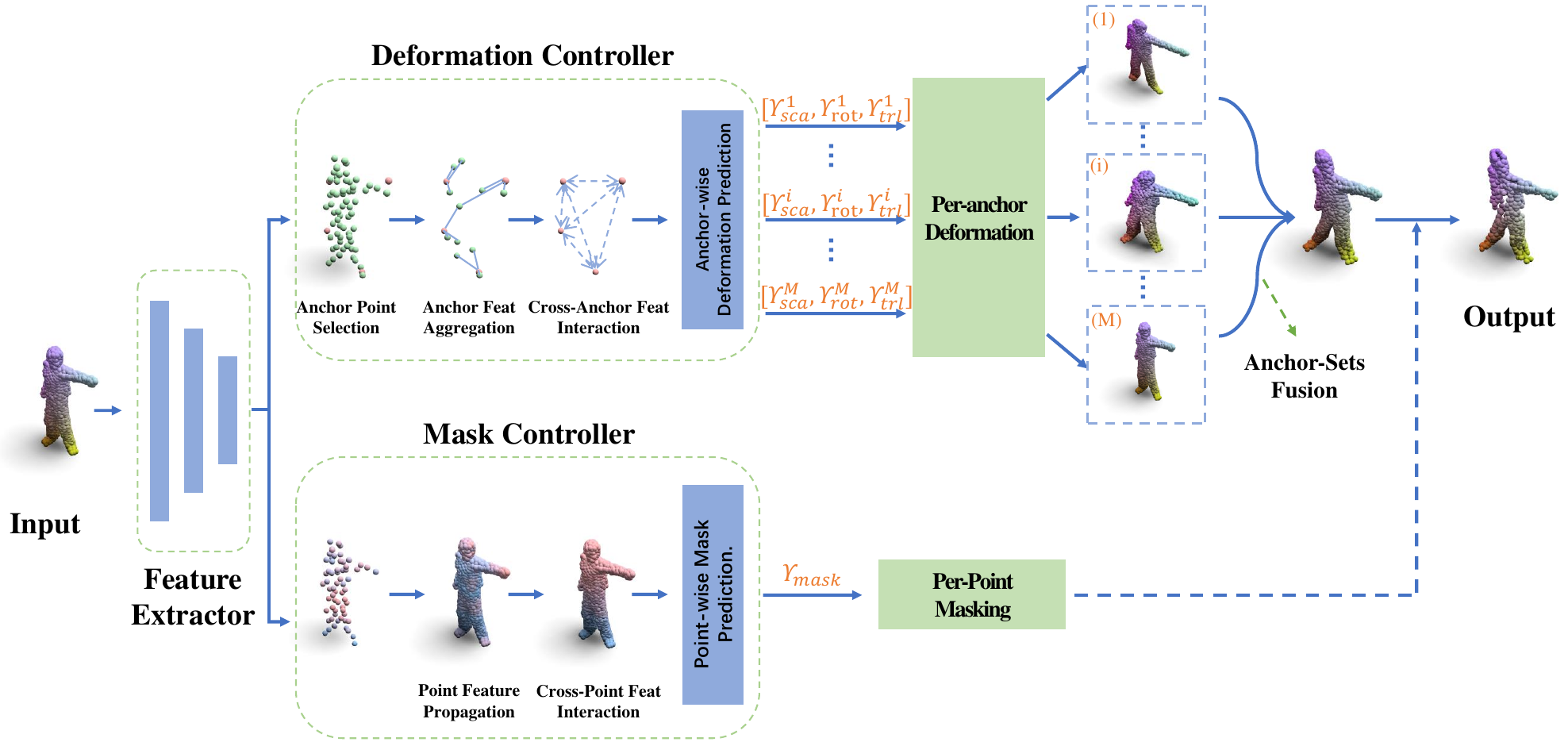}
\caption{\textbf{The overall architecture of sample-adaptive Imitator.} 
It initially leverages a Feature Extractor to extract local geometric features from the input point cloud. Subsequently, the Deformation Controller and Mask Controller modules are employed to predict deformation parameters specific to multiple anchors and generate a mask for the entire point cloud, respectively. A Corruption Simulator is then employed to apply corresponding transformations and masks to the input point cloud, resulting in an augmented point cloud.
}
\label{overview of Augmentor}
\vspace{-3mm}
\end{figure*}

\noindent\textbf{2D and 3D Robustness Benchmarks.}
Deep neural networks are known to be susceptible to corruption in data. To improve the robustness of these networks, ImageNet-C~\cite{hendrycks2019benchmarking_imagenetc} was introduced, which corrupts the test set of ImageNet~\cite{deng2009imagenet} using simulated corruptions such as motion blur. Similarly, two benchmarks for robust point cloud recognition under corruptions were introduced with ModelNet-C~\cite{ren2022benchmarking} and ShapeNet-C~\cite{ren2022pointcloud}. 
However, these two datasets are largely limited to CAD models, which fail to capture the variety of real-world corruptions encountered in practice and may hinder research progress in point cloud robustness. To bridge this gap, we present a novel point cloud corruption dataset, namely ScanObjectNN-C.


\vspace{-1mm}
\section{Method}
Given an input point cloud $\mathcal{P} \in \mathbb{R}^{N\times 3}$, our AdaptPoint aims to produce a corresponding corrupted point cloud $\mathcal{P}' \in \mathbb{R}^{N\times 3}$ based on the structural characteristics of $\mathcal{P}$ for improving the robustness of the classifier against corruptions. 
This is made possible through introducing a corruption imitator, which is jointly optimized with a discriminator alongside the classifier. Specifically, the imitator simulates common operations of corruptions in most real-life scenarios to ensure the plausibility. The discriminator helps weaken the generation of excessive corruption which is far away from raw data distribution. The classifier provides instant feedback and guide imitator to generate samples with suitable difficulty. 
These three key components are integrated into a self-contained framework and co-evolve as the training proceeds. With raw clean data as input, AdaptPoint progressively generates and refines corresponding corrupted data and robust classifier, which in turn reinforces all components in the loop. 


\subsection{Sample-adaptive Imitator}
\label{Imitator architecture}
The proposed sample-adaptive imitator workflow is presented in Fig.~\ref{overview of Augmentor}, consisting of four key modules: i) the Feature Extractor, implemented by PointNet++, captures local geometric features from the input point cloud $\mathcal{P}$, and outputs $N'$ sampled points $\mathcal{G} \in \mathbb{R}^{N'\times 3}$ with corresponding features $F = \left\{ f_j | j = 1,...,N' \right \} \in \mathbb{R}^{N'\times C}$;
ii) the Deformation Controller that responsible for capturing the interactions between selected anchor points and predicting deformation parameters specific to each anchor point, (iii) the Mask Controller that takes into consideration the overall structural information of the point cloud and predicts a per-point mask, and (iv) the Corruption Simulator that utilizes the predicted parameters and mask to apply transformations and occlusions to the input point cloud. 


\vspace{-3mm}
\subsubsection{Deformation Controller}
The Deformation Controller takes point cloud $\mathcal{P}$ and extracted feature $F$ as input, captures region features and enables multiple region transformations, thus generating deformation for point cloud by a convex combination of multiple locally-defined transformations.
To this end, we first select $M$ anchor points from $\mathcal{P}$ and aggregate local features centered on these anchor points, then establish the inner connections between them, finally predict deformation parameters, specifically for scaling $\gamma_{sca}=\left\{ \gamma_{sca}^i | i = 1,...,M \right\} \in \mathbb{R}^{M\times3}$, rotation $\gamma_{rot}=\left\{ \gamma_{rot}^i | i = 1,...,M \right\} \in \mathbb{R}^{M\times3}$ and translation $\gamma_{trl}=\left\{ \gamma_{trl}^i | i = 1,...,M \right\} \in \mathbb{R}^{M\times3}$ . 




\noindent\textbf{Per-anchor Feature aggregation.}
In this step, a set of $M$ anchor points $\mathcal{D} \in \mathbb{R}^{M\times3} $ are initially selected from $\mathcal{P}$, using the Farthest Point Sampling (FPS) algorithm, serving as the basis for the subsequent anchor-wise deformation process, where each anchor is independently transformed. 
For the $i$-th anchor, we select its K-nearest neighbors from $\mathcal{G}$ and learns its local feature representation:
\begin{equation}
    h_{i} = \underset{j =1,\cdots, K}{\mathcal{A}}\left\{\Phi\left(f_{i,j}\right)\right\}  \in \mathbb{R}^{1\times C},
\end{equation}
where $f_{i,j}$ is the $j$-th neighbor point feature of $i$-th anchor point. $\Phi(\cdot)$ is implemented by multi-layer perceptron (MLP). $\mathcal{A}(\cdot)$ denotes a symmetric function, \textit{e.g.}, max pooling, to aggregate encoded point features. As a result, anchor features $ H = \left \{ h_i | i = 1,...,M \right \} \in \mathbb{R}^{M\times C}$ are obtained.

\noindent\textbf{Cross-anchor Feature Interaction.}
\RDB{In pursuit of maintaining the cohesion of the comprehensive deformation, the mutual correlation of local transformations corresponding to each anchor set holds substantial relevance. A juxtaposition of myriad disconnected transformations may instigate a considerable disruption to the input. To counteract such unwelcome implications, it is essential to highlight the integration of anchor sets, along with building their connection.}

Therefore, a transformer network with multi-head-attention is utilized to maintain structural information of anchor sets.
Given anchor features $H$ with anchor coordinates $\mathcal{D}$, we first get query ($Q$), key ($K$) and value ($V$) as: 
\begin{equation}
     Q, K, V = H {W}^{Q},\ H {W}^{K},\ H {W}^{V},
\end{equation}
where $ {W}^{Q}, {W}^{K}, {W}^{V} \in \mathbb{R}^{C \times C}$ are learnable linear projections.
Next, we obtain updated anchor features $H' = \left \{ h_i' | i = 1,...,M \right \} \in \mathbb{R}^{M \times C}$ as:
\begin{equation}
    H' = MHA(Q, K, V, PE(\mathcal{D})),
\end{equation}
where $MHA$ represents a multi-head attention~\cite{vaswani2017attention} and $PE(\cdot)$ denotes the position embedding.

\noindent\textbf{Anchor-wise Deformation Prediction.}
Predicting deformation requires not only connection between anchor sets, but also global information to complement and constrain the overall shape of the object. 
Hence, the global structural information of all the anchors is encoded as 
\begin{equation}
    g = \underset{i =1,\cdots, M}{\mathcal{A}}\left\{\phi\left(h_i\right)\right\} \in \mathbb{R}^{1\times C},
\end{equation}
where $\phi$ is a point-wise feature embedding and $\mathcal{A}$ denotes max pooling, respectively.

The subsequent step involves predicting deformation parameters for each anchor point.
This is achieved by employing multiple linear layers with different activation functions for each deformation type:
\begin{equation}
\begin{split}
     \gamma_k^j = \alpha_k(\psi_k([h_j', g])), \ k\in \{sca, rot, trl\},
\end{split}
\end{equation}
where $[\cdot, \cdot]$ denotes the concatenation operation. $\alpha$ is the activation function and the mapping function $\psi$ is a point-wise feature transformations, such as linear projections or MLPs.
Specifically, Sigmoid activation function is used for scaling and Tanh activation function for rotation and translation. 

\vspace{-3mm}
\subsubsection{Mask Controller}



Occurrence of corruption leads to alterations not only in the geometric structure of the point cloud, but also results in partial omissions of certain parts to some extent. 
The Mask Controller accounts for the overall structure of the input point cloud and the interactions between all points, further predicting a point-wise mask to emulate the absence of points in the real world.
The presented module operates by initially getting point-wise feature at raw resolution,  $E \in \mathbb{R}^{N\times C}$, from $F \in \mathbb{R}^{N'\times C}$ through simple trilinear interpolation. 
It subsequently performs feature interaction across distinct points. The final output of this module is a point-wise mask $\gamma_{mask} = \left \{ \gamma_{mask}^j | j = 1,...,N \right \} \in \mathbb{R}^{N \times 1}$.

\noindent\textbf{Cross-point Feature Interaction.}
For improved point-wise mask prediction and mitigation of large-scale omissions that may distort the output samples, an approach that accounts for point-to-point connectivity is essential. We employ multi-head attention, similar as that used in the deformation controller, to model the relationships between points and get the updated point features $E' = \left \{ E_j' | j = 1,...,N\right \} \in \mathbb{R}^{N\times C}$.


\noindent\textbf{Point-wise Mask Prediction}. 
The accurate prediction of a point-wise Mask necessitates the utilization of both inter-point characteristics and holistic object structure. To this end, we first extract a global point feature $z \in \mathbb{R}^{1 \times C}$ by following the similar process as used in the deformation controller. 
Then we utilize the complementary local and global point features to predict the parameters of the mask:
\begin{equation}
    \gamma_{mask}^j = \sigma(\rho([E_j', z])) \in \mathbb{R}^{1 \times 1},
\end{equation}
where $\rho$ denotes a point-wise feature embedding learned by a MLP. $\sigma$ represents a normalization function implemented by gumbel-softmax. 

\vspace{-3mm}
\subsubsection{Corruption Simulator}
This module utilizes the predicted deformation parameters $\left\{\gamma_{sca}, \gamma_{rot}, \gamma_{trl} \right\}$ and mask $\gamma_{mask}$ to perform transformations input on point cloud.  
The resultant point cloud exhibits structural modifications and partial point loss, effectively simulating the characteristics of real-world corrupted point clouds.


\noindent\textbf{Per-anchor deformation.} 
To introduce local deformations in point clouds, we utilize a combination of scaling, rotation, and translation transformations with parameters predicted by the Deformation Controller module. 
Firstly, we normalize the input point cloud $\mathcal{P}$ using each anchor points in $\mathcal{D}$ to obtain a set of normalized point clouds $\mathcal{Q} = \{q_{i} \in \mathbb{R}^{ N \times 3} | i = 1,...,M \} \in \mathbb{R}^{M \times N \times 3} $.
Then, for the $i$-th normalized point cloud $q_i$, we apply a transformation based on the predicted deformation parameters, resulting in a deformed point cloud set $\mathcal{Q}' = \left \{ q_i' \in \mathbb{R}^{N \times 3} | i=1,...,M \right \} \in \mathbb{R}^{M \times N \times 3}$. Specifically, we apply the scaling, rotation, and translation transformations as follows:
\begin{equation}
q_i' = (q_i \cdot \gamma_{sca}^{i} ) R(\gamma_{rot}^{i}) + \gamma_{trl}^{i},
\end{equation}
where $R(\cdot)$ represents the rotation matrix. This process results in diverse local transformations for the deformed point cloud sets, with each set corresponding to a anchor.


\noindent\textbf{Anchor-Sets Fusion}.
We assemble the set of locally deformed point clouds $\mathcal{Q}'$ into a whole point cloud $\overline{\mathcal{P}} \in \mathbb{R}^{N \times 3}$  through anchor-sets fusion strategy~\cite{kim2021point}. 
To smoothly interpolate the local transformations in the 3D space, we employ the Nadaraya-Watson kernel regression inspired by the operation of \textit{Smooth deformations}~\cite{kim2021point}. 

\noindent\textbf{Per-point Masking}.
This step involves applying the predicted mask $\gamma_{mask} \in \mathbb{R}$ to the fused point cloud $\overline{\mathcal{P}}$. 
The masked point cloud $\mathcal{P}'$ is obtained as 
\begin{equation}
\mathcal{P}' = \overline{\mathcal{P}}  \odot \gamma_{mask} \in \mathbb{R}^{N \times 3}.
\end{equation}
Here, $\odot$ denotes element-wise multiplication. The purpose of mask is to imitate the effect of sensor noise or data missing that can occur during data acquisition. The resulting $\mathcal{P}'$ is taken as the final output augmented point cloud. 


\subsection{Learning Objectives}
\label{Loss of Imitator} 
We present the AdaptPoint framework that leverages a sample-adaptive imitator along with a discriminator and classifier. The loss function comprises of the discriminator loss, $\mathcal{L}_{adv}$, and the feedback loss, $\mathcal{L}_{feed}$, combined to form the complete loss function:
\begin{equation}
    \mathcal{L} = \mathcal{L}_{adv} + \lambda \mathcal{L}_{feed},
\end{equation}
where $\lambda$ is a balancing weight, set as $1$ in our experiments.

\noindent\textbf{Feedback Loss.}
To ensure the augmented data utilized for model training is more arduous than the original data yet not impractical due to excessive difficulty,  inspired by PoseAug~\cite{gong2021poseaug}, we introduce a feedback loss function. 
The feedback loss is designed to constrain the difference between the classification loss on augmented and original data within a proper range as follows:
\begin{equation}
\mathcal{L}_{feed} = |1 - exp[\mathcal{L_C}(\mathcal{P}') - \beta\mathcal{L_C}(
\mathcal{P})]|.
\end{equation}
where $\mathcal{L_C}$, the cross-entropy loss 
between predicted labels and ground truths, is used as the classifier loss metric.
Here, the parameter $\beta > 1$ controls the difficulty level of augmentation, ensuring that the value of $\mathcal{L_C}(\mathcal{P}')$ stays within a certain range relative to $\mathcal{L_C}(\mathcal{P})$. As the classifier's ability improves over time, we gradually increase the value of $\beta$ to elevate the degree of augmentation.

\noindent\textbf{Discriminator loss.}
\RDB{Merely striving for augmentations that maximize errors may spawn unfeasible training instances that infringe upon the structure of the point cloud, consequently inflicting detriment upon model performance. To alleviate such a concern, we incorporate a point cloud discriminator module within the point cloud's structure to facilitate the training of the augmentor, thereby confirming the plausibility of the augmented samples without undermining the diversity.}
Specially, we adopt PointNet++\cite{qi2017pointnet++} as the discriminator.
The adversarial loss is:
\begin{equation}
    \mathcal{L}_{adv} = \mathbb{E}_{x \sim \mathbb{P} (\mathcal{P})} [log(D(I(x)))],
\end{equation}
where $x$ is sampled from the probability distribution of the input point cloud data $\mathcal{P}$. $I(\cdot)$ denotes the mapping function learned by the imitator.
$D(I(x))$ is the discriminator's estimate of the probability that the augmented data sample $I(x)$ is clean.

\begin{figure}[!t]
\centering
\includegraphics[width= 0.5\textwidth]{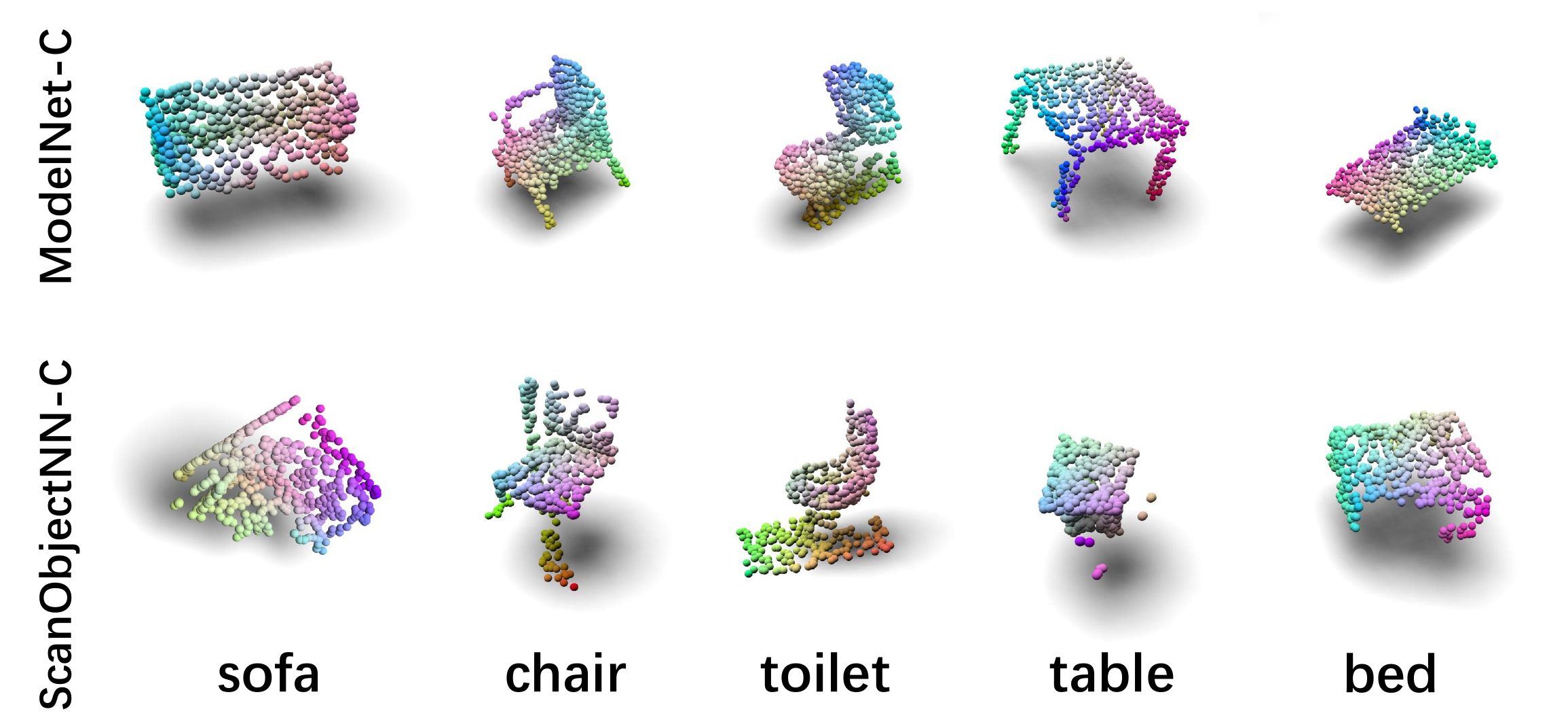} 
\caption{Some samples in ModelNet-C and ScanObjectNN-C.}
\label{comparison_modelnetc_scanobjectnnc}
\vspace{-6mm}
\end{figure}

\section{ScanObjectNN-C Dataset}
Current point cloud corruption ModelNet-C~\cite{ren2022benchmarking} and ShapeNet-C~\cite{ren2022pointcloud} are both from current point cloud percetion dataset~\cite{modelnet40, chang2015shapenet}, where the objects in that are simulated rather than scanned from real-world objects, may not adequately reflect a classifier's ability in realistic scenarios. 

To evaluate real-world robustness, We build a point cloud corruption dataset, dubbed as ScanObjectNN-C, which is more in line with the real-world corruption.
We collect point cloud from the test set of the most challenging variant of ScanObjectNN~\cite{uy2019revisiting}, PB-T50-RS, then process them by seven types of corruptions, “Jitter”, “Drop Global/Local”, “Add Global/Local”, “Scale” and “Rotate”, with five levels of severity to provide a comprehensive robustness evaluation. 
Specifically, objects from the ModelNet-C are derived from CAD models, whereas samples from ScanObjectNN-C are captured from real-world environments, often exhibiting occlusions, partiality, and complex backgrounds. 
The presented Fig.~\ref{comparison_modelnetc_scanobjectnnc} evidences the distinction between objects from the ModelNet-C dataset and samples from ScanObjectNN-C, indicating that the corrupting influence present in the ScanObjectNN-C more closely approximates the realistic conditions encountered in the real world, in comparison to ModelNet-C.

\section{Experiments}
We evaluate AdaptPoint on several point cloud corruption datasets and also test it for point cloud attack defense. 




\begin{table}[!t]
    \centering
    \setlength{\tabcolsep}{1pt}
    \caption{Classification results of mCE$(\%)$ on ModelNet-C.}
    \label{tab:modelnet40-c, mCE}
    \footnotesize
    \begin{tabular}{l|c|cccccccc}
        \toprule 
        Method  & mCE($\downarrow$) &  Sca & Jit & Drop-G & Drop-L & Add-G & Add-L & Rot \\
        \midrule
        DGCNN~\cite{wang2019dynamic} & 100.0 & 100.0 & 100.0 & 100.0 & 100.0 & 100.0 & 100.0 & 100.0 \\
        PointNet~\cite{qi2017pointnet}  & 142.2 & 126.6 & \textbf{64.2} & 50.0 & 107.2 & 298.0 & 159.3 & 190.2 \\
        RSCNN~\cite{liu2019relation}  & 113.0 & 107.4 & 117.1 & 80.6 & 151.7 & 71.2 & 115.3 & 147.9 \\
        SimpleView~\cite{goyal2021revisiting}  & 104.7 & 87.2 & 71.5 & 124.2 & 135.7 & 98.3 & 84.4 & 131.6 \\
        GDANet~\cite{xu2021learning}  & 89.2 & 83.0 & 83.9 & 79.4 & 89.4 & 87.1 & 103.6 & 98.1 \\
        CurveNet~\cite{xiang2021walk}  & 92.7 & 87.2 & 72.5 & 71.0 & 102.4 & 134.6 & 100.0 & 80.9 \\
        PAConv~\cite{xu2021paconv}  & 110.4 & 90.4 & 146.5 & 100.0 & 100.5 & 108.5 & 129.8 & 96.7 \\
        \hline
        PointNet++~\cite{qi2017pointnet++}  & 107.2 & 87.2 & 117.7 & 64.1 & 180.2 & 61.4 & 99.3 & 140.5 \\
        + PointWolf~\cite{kim2021point}  & 82.5 & \textbf{81.9} & 135.1 & 67.3 & 130.4 & 43.1 & 68.4 & 51.2 \\ 
        + Rsmix~\cite{lee2021regularization} & 86.3 & 89.4 & 164.9 & 48.4 & 73.9 & 26.1 & 32.7 & 168.4 \\ 
        + Wolfmix~\cite{ren2022benchmarking}  & 64.1 & 94.7 & 137.0 & 46.0 & 61.4 & 29.8 & 29.5 & 50.7 \\ 
        \rowcolor{RowColor} + AdaptPoint  & 63.7 & 109.6 & 102.2 & 36.7 & 66.7 & 30.5 & 40.0 & 60.5 \\ 
        \hline
        PointNeXt~\cite{qianpointnext} & 85.6  & 90.4  & 129.7  & 84.7  & 95.7  & 25.1  & 27.6  & 146.0  \\ 
        + PointWolf~\cite{kim2021point}   & 79.5  & 88.3  & 144.9  & 113.7  & 103.9  & 26.4  & 28.0  & 51.2  \\
        + Rsmix~\cite{lee2021regularization} & 87.9  & 100.0  & 159.2  & 86.7  & 54.1  & \textbf{23.1}  & 26.5  & 165.6  \\
        + Wolfmix~\cite{ren2022benchmarking}  & 74.0  & 84.0  & 156.0  & 119.4  & 56.5  & 23.7  & \textbf{25.1}  & 53.0  \\
        \rowcolor{RowColor} + AdaptPoint  & 67.7  & 101.1  & 116.8  & 60.5  & 69.1  & 29.5  & 32.4  & 64.2 \\ 
        \hline
        RPC~\cite{ren2022benchmarking}  & 86.3 & 84.0 & 89.2 & 49.2 & 79.7 & 92.9 & 101.1 & 107.9 \\
        +PointWolf~\cite{kim2021point}  & 70.2 & 88.3 & 107.3 & 80.2 & 104.8 & 27.1 & 29.1 & 54.4 \\ 
        +Rsmix~\cite{lee2021regularization} & 71.0 & 98.9 & 100.9 & 72.2 & 62.3 & 26.4 & 29.1 & 107.4 \\ 
        +Wolfmix~\cite{ren2022benchmarking}  & 60.1 & 101.1 & 96.8 & 42.3 & \textbf{51.2} & 33.2 & 48.0 & \textbf{47.9}\\ 
        \rowcolor{RowColor}+AdaptPoint  & \textbf{56.5} & 93.6 & 77.5 & \textbf{36.3} & 72.9 & 27.5 & 29.1 & 58.6 \\ 
        \bottomrule
    \end{tabular}
\vspace{-5mm}
\end{table}

\subsection{Results on ModelNet-C}
\label{sec: Results on ModelNet-C}

\noindent\textbf{Data and Setup.}
We train models on the clean ModelNet40~\cite{modelnet40} dataset and evaluate them on the ModelNet-C~\cite{ren2022benchmarking} corruption test suite. 
The imitator and discriminator are optimized using the Adam optimizer~\cite{kingma2014adam} and learning rates of 0.0001 and 0.0004, respectively. 
Mean Corruption Error metric (\textcolor{citegrey}{mCE, $\%, \downarrow$}) is taken as the main evaluation matrix.
More details are in the supplementary materials.

\noindent\textbf{Results.}
\RDB{In our experiments, PointNet++~\cite{qi2017pointnet++} and RPC~\cite{ren2022benchmarking} are adopted as the baseline models.}
Tab.~\ref{tab:modelnet40-c, mCE} presents the results of various methods on the robustness against corruption. Our proposed AdaptPoint outperforms all other methods and achieves the state-of-the-art mCE. The incorporation of AdaptPoint brings substantial improvements in model performance, as both PointNet++~\cite{qi2017pointnet++} and RPC~\cite{ren2022benchmarking} equipped with our method achieve an mCE of 63.7\% and 56.5\%, respectively, surpassing models without AdaptPoint by a large margin of over 20\%. 
Notably, our method improves the performance of nearly all categories, and exhibits the most significant improvements in drop-local and add-local, which evaluates the effectiveness of local deformation. 
Compared with PointWOLF~\cite{kim2021point} and Rsmix~\cite{lee2021regularization}, our method achieves a significant increase in mCE for jitter, over 30\% \RDB{on both baselines},
\RDB{which bears testament to AdaptPoint's efficacy in amplifying the local geometric nuances of the point cloud, thereby effectuating minor yet consequential modifications to the local structure.}

\begin{figure}[!t]
\centering
\includegraphics[width= 0.4\textwidth]{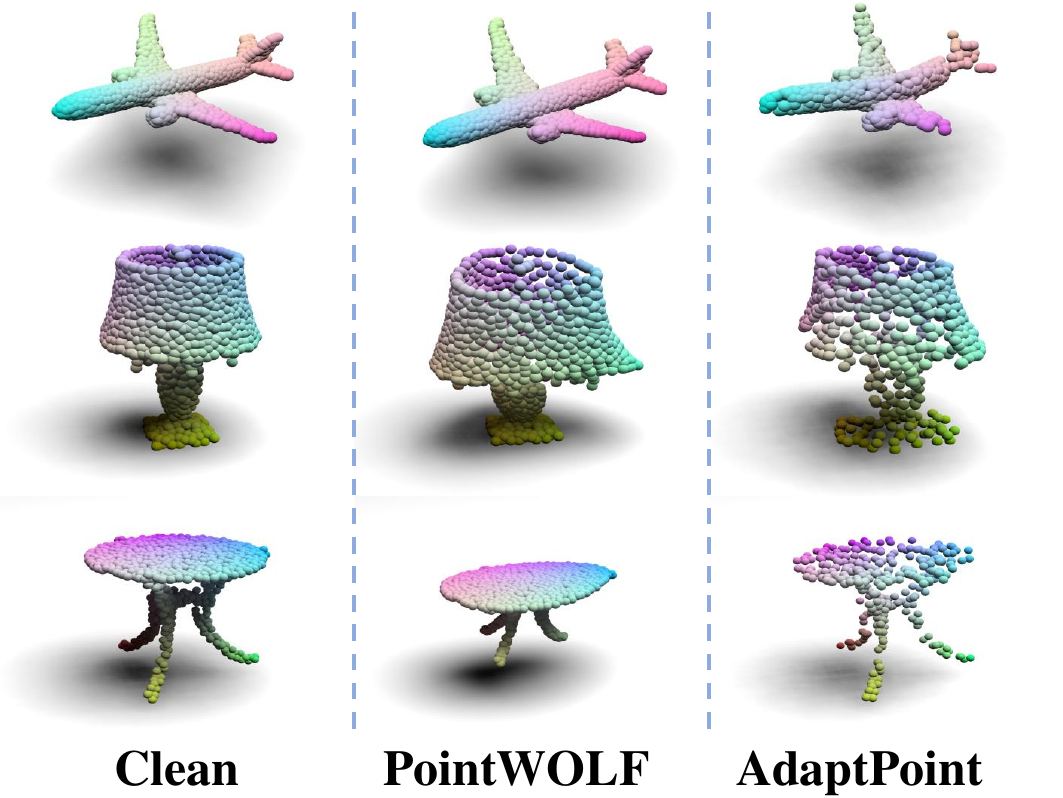} 
\caption{Examples of augmented samples by AdaptPoint. 
}
\label{Generated samples}
\vspace{-3mm}
\end{figure}


\noindent\textbf{Visualization of Augmented Samples.}
Fig.~\ref{Generated samples} showcases the visualization of both clean and augmented data generated by two distinct techniques, PointWOLF and AdaptPoint. This figure reveals that the augmented samples generated by AdaptPoint exhibit superior diversity compared to those generated by PointWOLF, thereby aligning better with real-world scenarios of corruption. 
\RDB{This signifies that our model induces a more intricate transformation in the geometry of the point cloud beyond just elementary distortions or absences. This advanced level of geometric modulation imparts robustness to the classifier model, as substantiated by the perceived enhancements in the quality of the produced samples.}

\begin{figure}[t!]
\centering
\includegraphics[width=0.45\textwidth]{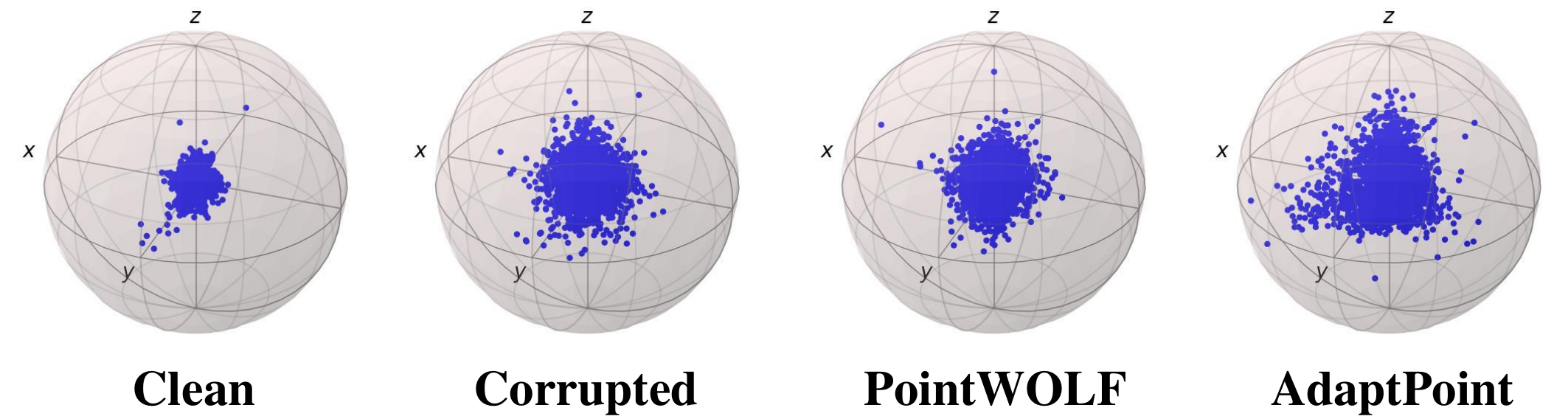}
\caption{Visualization of data distribution of raw data from ModelNet40, corrupted data by a commonly employed corruption technique (i.e., Drop-Global), and augmented data from PointWOLF and AdaptPoint. Our AdaptPoint outperforms the other methods in terms of generating diverse results. }
\label{Distribution Comparison}
\vspace{-6mm}
\end{figure}

\noindent\textbf{Visualization of Data Distribution.} 
To showcase the efficacy of AdaptPoint in augmenting data diversity, we conduct an analysis on the distribution of point cloud positions. As illustrated in Fig.~\ref{Distribution Comparison}, we compare the point position distributions of ModelNet40 with data corrupted by a commonly used corruption,(i.e., Drop-Global),  data augmented by\cite{kim2021point} and our method. Our findings reveal that the distribution in the original ModelNet40 dataset are concentrated near the origin of coordinates and have a limited range, which may explain why the models trained on this dataset exhibit poor generalization to corrupted scenarios. Furthermore, we observe that the point position distribution of the augmented data generated by~\cite{kim2021point} has a limited degree of divergence, indicating that the improvements made through the use of handcrafted rules are constrained. In contrast, our AdaptPoint method leverages a learnable augmentor to generate more diverse and realistic point positions, resulting in a greater degree of data augmentation.

\begin{figure}[!t]
\centering
\includegraphics[width= 0.5\textwidth]{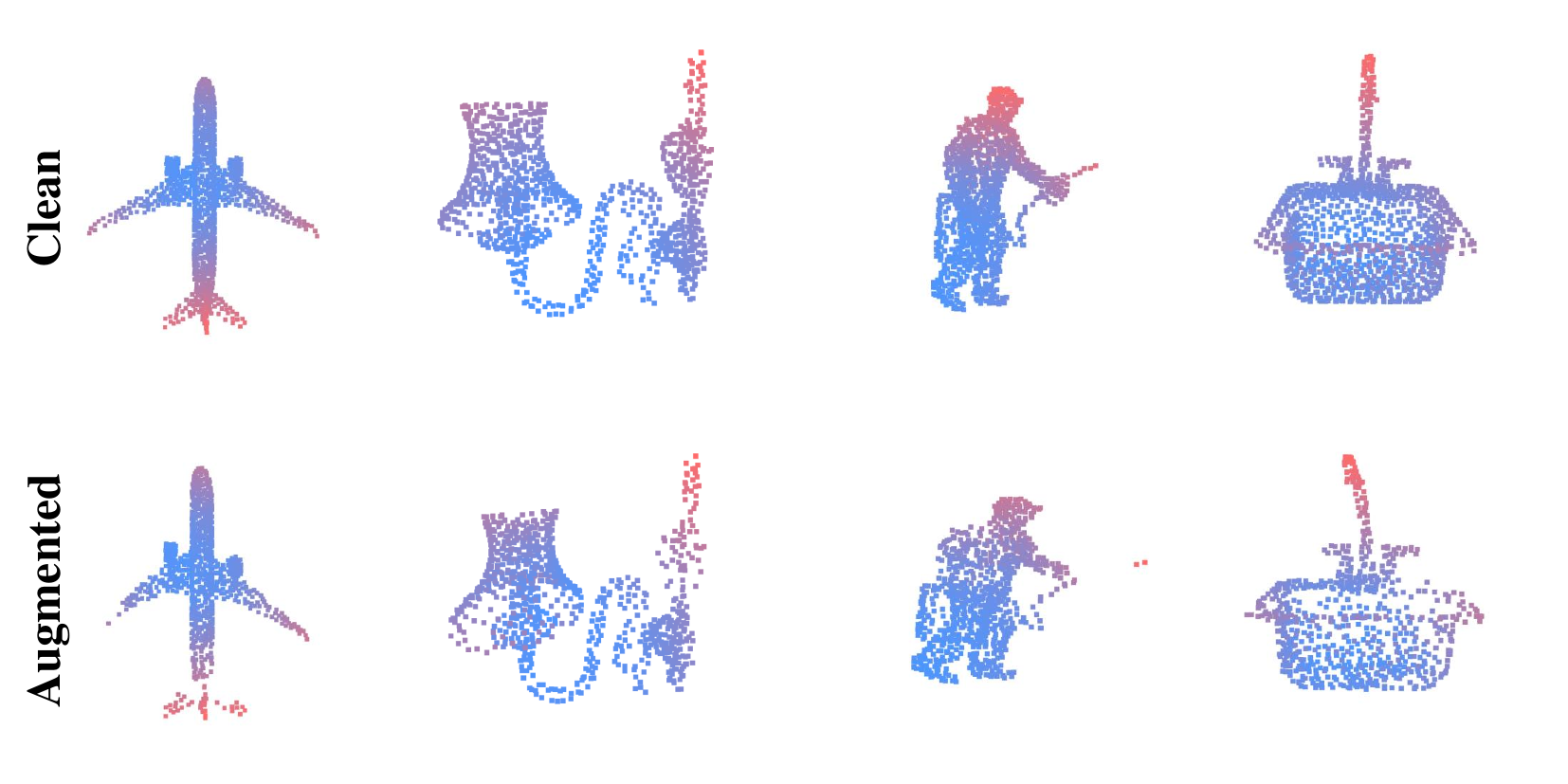} 
\caption{Visualization of salient geometry learned by the classifier. 
Points with high / low scores are in \textcolor{red}{red} / \textcolor{blue}{blue}.
}
\label{ablation study: attendweights}
\vspace{-6.5mm}
\end{figure}

\noindent\textbf{Visualization of Salient Geometry.}
In order to gain deeper insights into the impact of the imitator on clean data, we conducted a comparative analysis of the features 
attended to by the classifier on clean and augmented data. Fig.~\ref{ablation study: attendweights} provides a visual representation of the focus of the classifier's first perception layer. We computed the average features of the points and utilized them to represent their contribution to the classifier. 
Points that contributed more to the classifier were assigned higher scores in color, with red indicating high scores and blue indicating low scores. 
\RDB{The illustration exposes that, despite significant transformations in the morphology of the augmented data, it still maintains the crucial local prototypes and continues to supply the perception model with equivalent semantic information. }

\subsection{Results on ScanObjectNN-C}
\label{sec: Results on ScanObjectNN-C}
\noindent\textbf{Data and Setup.}
Objects in ScanObjectNN are scanned from real-world objects, and performance on ScanObjectNN may better reflect classifier ability in realistic scenarios. We train models on ScanObjectNN dataset and evaluate them on ScanObjectNN-C corruption test suite. 
We train the model for 250 epochs using the Adam optimizer with a learning rate of 0.002.

\noindent\textbf{Results.} 
Tab.~\ref{tab:scanobjectnn-c} presents the experimental results of our proposed method. Our method achieves mCE of 78.3\% on PointNeXt-S, outperforming the current state-of-the-art method WOLFMix~\cite{ren2022benchmarking}.
Across all models evaluated, our method achieved a lift of approximately 18\% in average mCE performance, which significantly surpassed the previous state-of-the-art method's improvement of 5.6\% in WOLFMix. 
Meanwhile, our method achieves overall accuracy (OA) of 88.45\% on ScanObjectNN, which is a highly competitive result surpassing many advanced methods~\cite{qianpointnext, ma2021rethinking}. 
These results demonstrate the generalizability of our proposed method and its potential for improving the performance of various models.

\begin{table}[!t]
    \centering
    \setlength{\tabcolsep}{0.5pt}
    \caption{Classification results of mCE$(\%)$ on ScanObjectNN-C.}
    \label{tab:scanobjectnn-c}
    \footnotesize
    \begin{tabular}{l|c|c|ccccccc}
        \toprule 
        Method & OA($\uparrow$)  & mCE($\downarrow$) & Sca & Jit & D-G & D-L & A-G & A-L & Rot  \\ 
        \hline
        DGCNN~\cite{wang2019dynamic} & 85.8 & 100.0  & 100.0  & 100.0  & 100.0  & 100.0  & 100.0  & 100.0  & 100.0  \\ 
        +PointWOLF & 85.6  & 99.6  & 89.5  & 104.6  & 104.1  & 98.3  & 98.0  & 100.9  & 101.6  \\ 
        +RSMix & 86.5  & 96.9  & 103.1  & 97.4  & 91.2  & 85.3  & 104.8  & 98.7  & 97.6  \\ 
        +WOLFMix & 87.2   & 92.3  & 92.1  & 102.6  & 92.6  & 85.3  & 96.2  & 85.7  & 91.5  \\ 
        \rowcolor{RowColor}+AdaptPoint & 84.4  & 90.2  & 90.6  & 107.5  & 72.3  & 73.0  & 93.3  & 93.3  & 101.4  \\ 
        \hline
        PointNet++~\cite{qi2017pointnet++} & 86.2 & 96.9  & 89.7  & 110.3  & 55.0  & 127.7  & 94.7  & 90.5  & 110.7  \\ 
        +PointWOLF & 86.6   & 96.4  & 84.0  & 108.7  & 70.5  & 156.6  & \textbf{87.7}  & 90.9  & 76.1  \\ 
        +RSMix & 87.3 & 91.9  & 89.0  & 100.7  & 55.6  & 99.0  & 94.6  & 89.1  & 114.9  \\ 
        +WOLFMix & 87.5  & 87.8  & \textbf{79.6}  & 109.0  & 64.2  & 117.7  & 88.1  & \textbf{79.8}  & 76.2  \\ 
        \rowcolor{RowColor}+AdaptPoint & 86.7  & 84.6  & 86.1  & 112.4  & 44.2  & 77.4  & 92.3  & 86.4  & 93.3  \\ 
        \hline
        RPC~\cite{ren2022benchmarking} & 74.7  & 132.6  & 131.7  & 107.3  & 145.5  & 130.5  & 114.2  & 158.7  & 140.2  \\ 
        +PointWOLF & 71.0   & 141.5  & 132.9  & 113.1  & 142.1  & 133.7  & 135.1  & 173.2  & 160.5  \\ 
        +RSMix & 65.7  & 146.9  & 149.3  & \textbf{94.8}  & 127.0  & 133.2  & 153.6  & 208.8  & 161.3  \\ 
        +WOLFMix & 79.7  & 120.2  & 108.5  & 117.4  & 122.3  & 107.2  & 123.3  & 124.6  & 138.1  \\ 
        \rowcolor{RowColor}+AdaptPoint & 83.5  & 96.5  & 106.4  & 100.5  & 66.9  & 77.8  & 107.3  & 105.9  & 110.9 \\ 
        \hline
        PointNeXt~\cite{qianpointnext} & 87.3   & 92.1  & 80.3  & 107.9  & 80.7  & 94.2  & 94.4  & 87.5  & 99.5  \\ 
        +PointWOLF & 87.4   & 89.5  & 81.4  & 112.9  & 89.8  & 92.3  & 95.0  & 83.7  & 71.1  \\ 
        +RSMix & 88.1   & 88.2  & 83.9  & 107.3  & 74.9  & 73.3  & 96.2  & 82.9  & 99.1  \\ 
        +WOLFMix & 87.7   & 86.9  & 81.9  & 119.3  & 89.7  & 78.0  & 89.3  & 80.0  & \textbf{70.0}  \\ 
        \rowcolor{RowColor}+AdaptPoint & \textbf{88.5}  & \textbf{78.3}  & 81.0  & 103.0  & \textbf{50.8}  & \textbf{62.8}  & 91.1  & 82.4  & 76.7  \\ 
        \bottomrule
    \end{tabular}
\vspace{-3mm}
\end{table}

\begin{table}[!t]
    \centering
    \setlength{\tabcolsep}{0.5pt}
    \caption{Segmenatation results of mCE $(\%)$ on ShapeNet-C}
    \label{tab:shapenet-c, mCE}
    \footnotesize
    \begin{tabular}{l|c|ccccccc}
        \toprule 
        Method & mCE($\downarrow$) & Sca & Jit & Drop-G & Drop-L & Add-G & Add-L & Rot \\ \hline
        DGCNN~\cite{wang2019dynamic} & 100.0  & 100.0  & 100.0  & 100.0  & 100.0  & 100.0  & 100.0  & 100.0  \\
        PointNet~\cite{qi2017pointnet} & 117.8  & 108.2  & 105.0  & 98.3  & 113.2  & 138.6  & 117.3  & 143.8  \\
        PointNet++~\cite{qi2017pointnet++} & 111.2  & 95.0  & 108.1  & 85.6  & 198.3  & 88.6  & 108.3  & 94.7  \\
        PAConv~\cite{xu2021paconv} & 92.7  & 92.7  & 107.2  & 92.5  & 92.7  & 74.3  & 94.8  & 94.8  \\ 
        GDANet~\cite{xu2021learning} & 92.3  & 92.2  & \textbf{101.2}  & 94.2  & 94.6  & 71.2  & 95.7  & 96.9  \\ 
        PT~\cite{zhao2021point} & 104.9  & 107.6  & 107.2  & 103.2  & 108.1  & 111.2  & 106.6  & 90.7  \\ 
        Point-MLP~\cite{ma2021rethinking} & 97.7  & 96.5  & 113.2  & 88.7  & 99.1  & 92.9  & 106.1  & 87.6  \\
        Point-BERT~\cite{yu2022point_pointbert} & 103.3  & 93.8  & 109.8  & 87.3  & 92.7  & 117.0  & 119.9  & 102.5  \\

        \midrule
        Point-MAE~\cite{pang2022masked} & 92.7  & \textbf{90.8}  & 103.5  & 85.2  & \textbf{88.2}  & 77.6  & 103.1  & 100.3  \\
        \rowcolor{RowColor}+AdaptPoint & \textbf{76.9}  & 95.2  & 104.8  & \textbf{84.5}  & 92.1  & \textbf{41.1}  & \textbf{42.6}  & \textbf{78.1} \\
        \bottomrule
    \end{tabular}
\vspace{-5mm}
\end{table}

\begin{table*}[!t]
\caption{Ablation studies }
\label{tab:ablation}
    \hspace{0mm}
    \subfloat[Effect of loss functions.]{
    \tabcolsep=1cm
    \tablestyle{6pt}{1.0}{
    \begin{tabular}{ccc}
    \toprule
    Feedback & Adv & mCE($\downarrow$) \\
    \hline
    \xmark & \cmark & 82.1 \\
    \cmark & \xmark & 83.2 \\
    \cmark & \cmark & \textbf{78.3} \\
    \bottomrule
    ~ & ~ & ~\\
    \end{tabular}
    } }
    \hspace{3mm}
    \subfloat[Effect of imitator components.]{
    \tabcolsep=0.3cm
    \tablestyle{6pt}{1.0}{
    \begin{tabular}{ccc}
    \toprule
    Deformation & Mask  & mCE($\downarrow$) \\
    \hline
    \xmark&\xmark & 92.1 \\
    \cmark&\xmark & 86.4 \\
    \xmark&\cmark & 87.1 \\  
    \cmark&\cmark & \textbf{78.3}  \\
    \bottomrule
    \end{tabular}
    }}
    \hspace{3mm}
    \subfloat[Effect of anchor number.]{
    \tabcolsep=1cm
    \footnotesize
    \tablestyle{14pt}{1.0}{
    \begin{tabular}{cc}
    \toprule
    Anchor & mCE($\downarrow$) \\
    \hline
    2 & 79.9 \\  
    4 & \textbf{78.3} \\
    8 & 80.2 \\    
    16 & 81.4\\    
    \bottomrule
    \end{tabular}
    }}
    \hspace{3mm}
    \subfloat[Impact of loss weight $\lambda$.]{
    \tabcolsep=1cm
    \footnotesize
    \tablestyle{14pt}{1.0}{
    \begin{tabular}{cc}
    \toprule
    $\lambda$  & mCE($\downarrow$) \\
    \hline
    0.5 & 78.7  \\  
    1 & \textbf{78.3} \\
    2 & 79.0  \\    
    \bottomrule
    ~ & ~ \\
    \end{tabular}
    }}  \\

\vspace{-6mm}
\end{table*}

\begin{table}[!t]
    \centering
    \setlength{\tabcolsep}{1.9pt}
    \caption{
    Results on point cloud attack defense (OA, $\%$).}

    \label{tab: attack results}
    \footnotesize
    \begin{tabular}{l|cccccc}
        \toprule 
        Method & Perturb & Add-CD & Add-HD & kNN & Drop-100 & Drop-200 \\ 
        \hline
        NoDefense & - & 7.24 & 6.59 & - & 80.19 & 68.96  \\
        SRS~\cite{yang2019adversarial} & 73.14 & 65.32 & 43.11 & 49.96 & 64.51 & 39.60 \\ 
        SOR~\cite{zhou2020lg} & 77.67 & 72.90 & 72.41 & 61.35 & 74.16 & 69.17 \\ 
        SOR-AE~\cite{zhou2020lg} & 78.73 & 73.38 & 71.19 & 78.73 & 76.66 & 68.23 \\ 
        Adv Training & 20.03 & 12.27 & 10.06 & 8.63 & 80.39 & 67.14 \\ 
        DUP-Net~\cite{zhou2019dup} & 80.63 & 75.81 & 72.45 & 74.88 & 76.38 & 72.00 \\ 
        IF-Defense~\cite{wu2020if} & \textbf{86.99} & 80.19 & 76.09 & \textbf{85.62} & 84.56 & 79.09 \\ 
        \hline
        \rowcolor{RowColor}AdaptPoint & 86.75 & \textbf{80.83} & \textbf{77.03} & 77.67 & \textbf{86.55} & \textbf{83.59} \\ 
        \bottomrule
    \end{tabular}
\vspace{-5mm}
\end{table}

\subsection{Results on Shapenet-C}
\noindent\textbf{Data and Setup.}
We conduct part segmentation experiment on ShapeNet-C~\cite{ren2022pointcloud}. We employed PointMAE~\cite{pang2022masked} as the baseline and trained the model for 300 epochs with the AdamW optimizer. The learning rate was set to 0.0002. 

\noindent\textbf{Results.}
Tab.~\ref{tab:shapenet-c, mCE} demonstrates that our method achieves a score of $76.9\%$ mCE and outperforms the state-of-the-art PointMAE method by a margin of 15.4\% . It is worth mentioning that PointMAE is a MAE model, and the significant improvement provided by our AdaptPoint suggests its effectiveness in enhancing the performance of self-supervised models, beyond being limited to fully supervised models. 
In addition to its efficacy in classification, our proposed AdaptPoint augmentation technique showcases its potential on ShapeNet-C dataset by achieving favorable performance. Remarkably, our method achieved these results while preserving the integrity of the semantic information, attesting to its robustness and practicality.
\RDB{demonstrating superiority of our approach in downstream tasks}


\subsection{Results on Point Cloud Attack Defense}
\noindent\textbf{Setup.}
Motivated by the remarkable performance of the AdaptPoint method against corruption, we explore its application in point cloud defense. We evaluate the performance of  PointNet++~\cite{qi2017pointnet++} trained using the AdaptPoint method on point perturbation attack~\cite{xiang2019generating}, individual point adding attack~\cite{xiang2019generating}, kNN attack~\cite{tsai2020robust} and point dropping attack~\cite{zheng2019pointcloud}.

\noindent\textbf{Results.}
Based on the results presented in Tab.~\ref{tab: attack results}, it is observed that our AdaptPoint consistently outperforms all other defense methods for two-point dropping and adding attacks. The accuracy improvement can be as high as 4.5\% in the Drop-200 attack, indicating the scalability and effectiveness of the approach. Even with the Perturb attack, which introduces very fine perturbations leading to a chaotic point cloud, AdaptPoint achieves the second-best performance, closely approaching the state-of-the-art defense method~\cite{wu2020if}. The above analysis leads to the conclusion that our method has a strong generalization ability across various point cloud attack algorithms.

\subsection{Ablation Studies}

PointNeXt~\cite{qianpointnext} is taken as the classifier in ablations. More implementation details and ablations on hyperparameters are available in supplementary materials.

\noindent\textbf{Effect of Feedback and Adversarial Loss.} 
To ascertain the impact of feedback loss, we conducted an ablation study by removing feedback loss. Tab.~\ref{tab:ablation}\textcolor{red}{a}, unequivocally demonstrate that removing feedback loss led to a dramatic drop ($4\%\downarrow$) in model performance, underscoring it critical importance. Besides, we also illustrate the effectiveness of plausibility guidance from the discriminator. The remove of Adversarial loss leads to $4.9\%$ performance drop, demonstrating the crucial role of point cloud plausibility in enhancing classifier performance, with the discriminator playing a pivotal role in the process.

\noindent\textbf{Effect of Deformation and Mask Controller.}
We conducted an investigation into the efficacy of the Deformation and Mask controller. Tab.~\ref{tab:ablation}\textcolor{red}{b} presents the results obtained by adding each component to a base pipeline. The performance gains achieved by incorporating each component alone were limited. Notably, combining both deformation and mask led to a substantial decrease in mCE, with the value reducing from 92.1\% to 78.3\%. This finding suggests that the incorporation of both deformation and mask significantly improves model robustness.


\noindent\textbf{Effect of the Number of Anchors.}
We investigate the impact of anchor numbers of the deformation controller. As illustrated in Tab.~\ref{tab:ablation}\textcolor{red}{c}, we vary the number of anchors and analyze the resulting mCE scores. The results show that the performance of mCE is 81.4\% when the number of weight matrices is 16, which is 3.1\% higher than the performance achieved using 4 anchors. This observation can be attributed to the fact that too many anchors can negatively impact the overall deformation, therefore leading to a decrease in performance.

\noindent\textbf{Impact of Feedback Loss Weight.}
Tab.~\ref{tab:ablation}\textcolor{red}{d} presents an empirical investigation of the impact of loss weight on the feedback loss in the context of AdaptPoint. The findings demonstrate that modifying the loss weight $\lambda$ can lead to a harmful effect on model performance, due to the disruption of the delicate balance between the adversarial loss and the feedback loss. Specifically, increasing or decreasing the loss weight results in a drop in performance. 
Thus, maintaining a proper balance among all components in AdaptPoint is crucial, which further corroborates the efficacy of our imitator loss design.

\noindent\textbf{Combination of corruption techniques.}
There exist two common techniques for point cloud data augmentation: predefined transformations like PointWOLF, and mix-up based sample generation like RSMix. {WOLFMix combines both techniques}, leveraging their strengths. 
Results in Tab.~\ref{tab:modelnet40-c, mCE} in Sec.~\ref{sec: Results on ModelNet-C} and Tab.~\ref{tab:scanobjectnn-c} in Sec.~\ref{sec: Results on ScanObjectNN-C}, involve all these techniques under both standalone and combined circumstances, offering a holistic comparison with our method.

\section{Conclusion}
In this paper, we present a novel auto-augmentation methodology, 
denominated AdaptPoint, 
specifically designed for point cloud recognition under real-world corruptions. The AdaptPoint approach leverages the benefits of a sample-adaptive imitator, enabling the simulation of real-world corruptions scenarios. Additionally, we introduce a real-world corruption dataset to facilitate the evaluation of point cloud recognition in the presence of real-world corruptions. Our experimental results, conducted on three benchmark datasets, manifest the superiority of our proposed approach.

\vspace{3mm}
\noindent\textbf{Acknowledgments}
This work was financially supported by the National Natural Science Foundation of China (No. 62101032), the Postdoctoral Science Foundation of China (Nos. 2021M690015, 2022T150050), and Beijing Institute of Technology Research Fund Program for Young Scholars (No. 3040011182111).

{\small
\bibliographystyle{ieee_fullname}
\bibliography{egbib}

\begin{thebibliography}{10}\itemsep=-1pt

\bibitem{chang2015shapenet}
Angel~X Chang, Thomas Funkhouser, Leonidas Guibas, Pat Hanrahan, Qixing Huang,
  Zimo Li, Silvio Savarese, Manolis Savva, Shuran Song, Hao Su, et~al.
\newblock Shapenet: An information-rich 3d model repository.
\newblock {\em arXiv preprint arXiv:1512.03012}, 2015.

\bibitem{chen2020pointmixup}
Yunlu Chen, Vincent~Tao Hu, Efstratios Gavves, Thomas Mensink, Pascal Mettes,
  Pengwan Yang, and Cees~GM Snoek.
\newblock Pointmixup: Augmentation for point clouds.
\newblock In {\em ECCV}, pages 330--345. Springer, 2020.

\bibitem{deng2009imagenet}
Jia Deng, Wei Dong, Richard Socher, Li-Jia Li, Kai Li, and Li Fei-Fei.
\newblock Imagenet: A large-scale hierarchical image database.
\newblock In {\em CVPR}, pages 248--255. Ieee, 2009.

\bibitem{gong2021poseaug}
Kehong Gong, Jianfeng Zhang, and Jiashi Feng.
\newblock Poseaug: A differentiable pose augmentation framework for 3d human
  pose estimation.
\newblock In {\em CVPR}, pages 8575--8584, 2021.

\bibitem{goyal2021revisiting}
Ankit Goyal, Hei Law, Bowei Liu, Alejandro Newell, and Jia Deng.
\newblock Revisiting point cloud shape classification with a simple and
  effective baseline.
\newblock In {\em ICML}, pages 3809--3820, 2021.

\bibitem{hendrycks2019benchmarking_imagenetc}
Dan Hendrycks and Thomas Dietterich.
\newblock Benchmarking neural network robustness to common corruptions and
  perturbations.
\newblock {\em arXiv preprint arXiv:1903.12261}, 2019.

\bibitem{kim2021point}
Sihyeon Kim, Sanghyeok Lee, Dasol Hwang, Jaewon Lee, Seong~Jae Hwang, and
  Hyunwoo~J Kim.
\newblock Point cloud augmentation with weighted local transformations.
\newblock In {\em ICCV}, pages 548--557, 2021.

\bibitem{kingma2014adam}
Diederik~P Kingma and Jimmy Ba.
\newblock Adam: A method for stochastic optimization.
\newblock {\em arXiv preprint arXiv:1412.6980}, 2014.

\bibitem{lee2021regularization}
Dogyoon Lee, Jaeha Lee, Junhyeop Lee, Hyeongmin Lee, Minhyeok Lee, Sungmin Woo,
  and Sangyoun Lee.
\newblock Regularization strategy for point cloud via rigidly mixed sample.
\newblock In {\em CVPR}, pages 15900--15909, 2021.

\bibitem{li2020pointaugment}
Ruihui Li, Xianzhi Li, Pheng-Ann Heng, and Chi-Wing Fu.
\newblock Pointaugment: an auto-augmentation framework for point cloud
  classification.
\newblock In {\em CVPR}, pages 6378--6387, 2020.

\bibitem{liu2019relation}
Yongcheng Liu, Bin Fan, Shiming Xiang, and Chunhong Pan.
\newblock Relation-shape convolutional neural network for point cloud analysis.
\newblock In {\em CVPR}, 2019.

\bibitem{ma2021rethinking}
Xu Ma, Can Qin, Haoxuan You, Haoxi Ran, and Yun Fu.
\newblock Rethinking network design and local geometry in point cloud: A simple
  residual mlp framework.
\newblock In {\em ICLR}, 2022.

\bibitem{pang2022masked}
Yatian Pang, Wenxiao Wang, Francis~EH Tay, Wei Liu, Yonghong Tian, and Li Yuan.
\newblock Masked autoencoders for point cloud self-supervised learning.
\newblock In {\em ECCV}, pages 604--621. Springer, 2022.

\bibitem{qi2017pointnet}
Charles~R Qi, Hao Su, Kaichun Mo, and Leonidas~J Guibas.
\newblock Pointnet: Deep learning on point sets for 3d classification and
  segmentation.
\newblock In {\em CVPR}, 2017.

\bibitem{qi2017pointnet++}
Charles~Ruizhongtai Qi, Li Yi, Hao Su, and Leonidas~J Guibas.
\newblock Pointnet++: Deep hierarchical feature learning on point sets in a
  metric space.
\newblock In {\em NIPS}, 2017.

\bibitem{qianpointnext}
Guocheng Qian, Yuchen Li, Houwen Peng, Jinjie Mai, Hasan Abed Al~Kader Hammoud,
  Mohamed Elhoseiny, and Bernard Ghanem.
\newblock Pointnext: Revisiting pointnet++ with improved training and scaling
  strategies.
\newblock In {\em NIPS}, 2022.

\bibitem{ren2022pointcloud}
Jiawei Ren, Lingdong Kong, Liang Pan, and Ziwei Liu.
\newblock Pointcloud-c: Benchmarking and analyzing point cloud perception
  robustness under corruptions.
\newblock {\em preprint}, 2022.

\bibitem{ren2022benchmarking}
Jiawei Ren, Liang Pan, and Ziwei Liu.
\newblock Benchmarking and analyzing point cloud classification under
  corruptions.
\newblock {\em arXiv preprint arXiv:2202.03377}, 2022.

\bibitem{tsai2020robust}
Tzungyu Tsai, Kaichen Yang, Tsung-Yi Ho, and Yier Jin.
\newblock Robust adversarial objects against deep learning models.
\newblock In {\em AAAI}, volume~34, pages 954--962, 2020.

\bibitem{uy2019revisiting}
Mikaela~Angelina Uy, Quang-Hieu Pham, Binh-Son Hua, Thanh Nguyen, and Sai-Kit
  Yeung.
\newblock Revisiting point cloud classification: A new benchmark dataset and
  classification model on real-world data.
\newblock In {\em ICCV}, 2019.

\bibitem{vaswani2017attention}
Ashish Vaswani, Noam Shazeer, Niki Parmar, Jakob Uszkoreit, Llion Jones,
  Aidan~N Gomez, {\L}ukasz Kaiser, and Illia Polosukhin.
\newblock Attention is all you need.
\newblock {\em NIPS}, 30, 2017.

\bibitem{wang2021papooling}
Jie Wang, Jianan Li, Lihe Ding, Ying Wang, and Tingfa Xu.
\newblock Papooling: Graph-based position adaptive aggregation of local
  geometry in point clouds.
\newblock {\em arXiv preprint arXiv:2111.14067}, 2021.

\bibitem{wang2019dynamic}
Yue Wang, Yongbin Sun, Ziwei Liu, Sanjay~E Sarma, Michael~M Bronstein, and
  Justin~M Solomon.
\newblock Dynamic graph cnn for learning on point clouds.
\newblock {\em TOG}, 38(5):1--12, 2019.

\bibitem{wu2019pointconv}
Wenxuan Wu, Zhongang Qi, and Li Fuxin.
\newblock Pointconv: Deep convolutional networks on 3d point clouds.
\newblock In {\em CVPR}, 2019.

\bibitem{wu2020if}
Ziyi Wu, Yueqi Duan, He Wang, Qingnan Fan, and Leonidas~J Guibas.
\newblock If-defense: 3d adversarial point cloud defense via implicit function
  based restoration.
\newblock {\em arXiv preprint arXiv:2010.05272}, 2020.

\bibitem{modelnet40}
Zhirong Wu, Shuran Song, Aditya Khosla, Fisher Yu, Linguang Zhang, Xiaoou Tang,
  and Jianxiong Xiao.
\newblock 3d shapenets: A deep representation for volumetric shapes.
\newblock In {\em CVPR}, 2015.

\bibitem{xiang2019generating}
Chong Xiang, Charles~R Qi, and Bo Li.
\newblock Generating 3d adversarial point clouds.
\newblock In {\em CVPR}, pages 9136--9144, 2019.

\bibitem{xiang2021walk}
Tiange Xiang, Chaoyi Zhang, Yang Song, Jianhui Yu, and Weidong Cai.
\newblock Walk in the cloud: Learning curves for point clouds shape analysis.
\newblock In {\em ICCV}, 2021.

\bibitem{xu2021paconv}
Mutian Xu, Runyu Ding, Hengshuang Zhao, and Xiaojuan Qi.
\newblock Paconv: Position adaptive convolution with dynamic kernel assembling
  on point clouds.
\newblock In {\em CVPR}, 2021.

\bibitem{xu2021learning}
Mutian Xu, Junhao Zhang, Zhipeng Zhou, Mingye Xu, Xiaojuan Qi, and Yu Qiao.
\newblock Learning geometry-disentangled representation for complementary
  understanding of 3d object point cloud.
\newblock In {\em AAAI}, volume~35, pages 3056--3064, 2021.

\bibitem{yang2019adversarial}
Jiancheng Yang, Qiang Zhang, Rongyao Fang, Bingbing Ni, Jinxian Liu, and Qi
  Tian.
\newblock Adversarial attack and defense on point sets.
\newblock {\em arXiv preprint arXiv:1902.10899}, 2019.

\bibitem{yu2022point_pointbert}
Xumin Yu, Lulu Tang, Yongming Rao, Tiejun Huang, Jie Zhou, and Jiwen Lu.
\newblock Point-bert: Pre-training 3d point cloud transformers with masked
  point modeling.
\newblock In {\em CVPR}, pages 19313--19322, 2022.

\bibitem{zhao2021point}
Hengshuang Zhao, Li Jiang, Jiaya Jia, Philip~HS Torr, and Vladlen Koltun.
\newblock Point transformer.
\newblock In {\em ICCV}, 2021.

\bibitem{zheng2019pointcloud}
Tianhang Zheng, Changyou Chen, Junsong Yuan, Bo Li, and Kui Ren.
\newblock Pointcloud saliency maps.
\newblock In {\em ICCV}, pages 1598--1606, 2019.

\bibitem{zhou2020lg}
Hang Zhou, Dongdong Chen, Jing Liao, Kejiang Chen, Xiaoyi Dong, Kunlin Liu,
  Weiming Zhang, Gang Hua, and Nenghai Yu.
\newblock Lg-gan: Label guided adversarial network for flexible targeted attack
  of point cloud based deep networks.
\newblock In {\em CVPR}, pages 10356--10365, 2020.

\bibitem{zhou2019dup}
Hang Zhou, Kejiang Chen, Weiming Zhang, Han Fang, Wenbo Zhou, and Nenghai Yu.
\newblock Dup-net: Denoiser and upsampler network for 3d adversarial point
  clouds defense.
\newblock In {\em CVPR}, pages 1961--1970, 2019.

\end{thebibliography}
}


\end{document}